\documentclass[letterpaper]{article} 
\usepackage{aaai2026}  
\nocopyright
\usepackage{times}  
\usepackage{helvet}  
\usepackage{courier}  
\usepackage[hyphens]{url}  
\usepackage{graphicx} 
\usepackage{natbib}  
\usepackage{caption} 
\usepackage{algorithm}
\usepackage{algorithmic}

\usepackage{amssymb}

\usepackage{newfloat}
\usepackage{listings}
\DeclareCaptionStyle{ruled}{labelfont=normalfont,labelsep=colon,strut=off} 
\floatstyle{ruled}
\newfloat{listing}{tb}{lst}{}
\floatname{listing}{Listing}
\title{Theory of Mind for Explainable Human-Robot Interaction}
\author{
   Marie S. Bauer, Julia Gachot, Matthias Kerzel, Cornelius Weber, Stefan Wermter\\
}
\affiliations{
    Knowledge Technology, Department of Informatics, University of Hamburg\\
    Vogt-Koelln-Str. 30\\
    22527 Hamburg, Germany\\
    \{marie.bauer, julia.gachot, matthias.kerzel, cornelius.weber, stefan.wermter\}@uni-hamburg.de
}

\usepackage{bibentry}

\begin{document}

\maketitle

\begin{abstract} 
Within the context of human–robot interaction (HRI), Theory of Mind (ToM) is intended to serve as a user-friendly backend to the interface of robotic systems, enabling robots to infer and respond to human mental states. 
When integrated into robots, ToM allows them to adapt their internal models to users’ behaviors, enhancing the interpretability and predictability of their actions.
Similarly, Explainable Artificial Intelligence (XAI) aims to make AI systems transparent and interpretable, allowing humans to understand and interact with them effectively.
Since ToM in HRI serves related purposes, we propose to consider ToM as a form of XAI and evaluate it through the eValuation XAI (VXAI) framework and its seven desiderata.
This paper identifies a critical gap in the application of ToM within HRI, as existing methods rarely assess the extent to which explanations correspond to the robot’s actual internal reasoning.
To address this limitation, we propose to integrate ToM within XAI frameworks.
By embedding ToM principles inside XAI, we argue for a shift in perspective, as current XAI research focuses predominantly on the AI system itself and often lacks user-centered explanations. 
Incorporating ToM would enable a change in focus, prioritizing the user’s informational needs and perspective.
\end{abstract}


\section{Introduction}
As interactions between humans and robots become increasingly common \cite{Lee-service-robot}, it is intuitive to seek more human-like modes of interaction to be able to understand robots' behaviors \cite{sridharan-2019, Kerzel2023}. 
This need naturally motivates the application of ToM in HRI. 
ToM refers to the human ability to attribute mental states such as beliefs, desires, and intentions to oneself and others to predict and explain behavior \cite{Premack_Woodruff_1978}.
When embedded in robots, ToM methods emphasize understanding and adapting to users’ mental states, and this can be used to produce explanations that are often more intuitive and user-friendly \cite{williams-2022}. 
ToM also allows robots to interpret and respond to users’ inferred mental states, fostering more natural, adaptive, and transparent interactions \cite{Yuan}. 
On the other hand, XAI aims to make black box models more transparent and interpretable; however, it frequently overlooks user-centered evaluations \cite{Rong}. 

Since both XAI and ToM in HRI aim to make internal reasoning more understandable to humans and enhance human–AI collaboration, we propose considering ToM as a form of XAI, and therefore evaluate it accordingly.
To this end, we evaluated recent ToM studies in HRI using an XAI evaluation framework and identified some limitations. 
Existing ToM approaches rarely assess whether the information presented to users accurately reflects the robot’s internal reasoning, nor do they evaluate the robustness and reproducibility of the explanations.

To address gaps in both ToM and XAI, particularly regarding explanation fidelity and user-centered evaluation, we propose leveraging ToM within an XAI framework, combining ToM’s user-centered perspective with XAI’s technical rigor. 
This shift in perspective aims to enable evaluations that encompass both fidelity to the model and alignment with user understanding, ultimately narrowing the gap between system transparency and human interpretability.

\section{Theory of Mind for HRI}
ToM is often treated as a heuristic in artificial intelligence, where one of the participants is replaced by a robot. 
In this section, we review recent studies that have used ToM in HRI to evaluate human-AI collaboration and understanding. 

\subsection{Attributing ToM to Agents}
Several studies examine whether humans naturally attribute ToM to robots even in the absence of explicit ToM mechanisms. 
A first study found that humans are able to interpret robots’ behavior similarly to human behavior, provided that the robots display distinct and interpretable social cues. 
However, when a robot’s cues deviate from human expectations, this understanding diminishes \cite{Bank}.
A second study, which examined the robustness and conviction of large language models (LLMs), demonstrated that while LLMs can serve as a useful tool in human–robot interaction \cite{Becker}, they do not function as reliable ToM agents \cite{Verma}.
These findings suggest that effective human–robot interaction is facilitated when robots produce responses that align with typical human behavior, and highlight the need for the integration of explicit ToM-like mechanisms in robotic systems.

\begin{table*}[ht]
\centering
\renewcommand{\arraystretch}{1.2} 
\fontsize{9}{11}\selectfont
\setlength{\tabcolsep}{1mm}
\begin{tabular}{|c|c|c|c|c|c|c|c|}
\hline
\textbf{Papers} & \textbf{Parsimony} & \textbf{Plausibility} & \textbf{Coverage} & \textbf{Fidelity} & \textbf{Continuity} & \textbf{Consistency} & \textbf{Efficiency} \\ \hline
\citet{Bank} & $\checkmark$& $\checkmark$&  &  &  & & $\checkmark$\\ \hline
\citet{Mou} & $\checkmark$& $\checkmark$&  &  &  & & \\ \hline
\citet{Cantucci} & $\checkmark$& $\checkmark$&  & &  & & $\checkmark$\\ \hline
\citet{Kerzel} & $\checkmark$& $\checkmark$&  & &  & & $\checkmark$\\ \hline
\citet{Shvo} & $\checkmark$& $\checkmark$&  &  &  & & $\checkmark$\\ \hline
\citet{Yuan} & $\checkmark$& $\checkmark$& & & $\checkmark$& $\checkmark$ & $\checkmark$\\ \hline
\citet{Verma} & $\checkmark$& $\checkmark$ &  &  &  &  & $\checkmark$\\ \hline
\citet{Angelopoulos} & $\checkmark$& $\checkmark$& & & $\checkmark$& $\checkmark$ & $\checkmark$\\ \hline
\end{tabular}
\caption{Evaluation of ToM in HRI studies using the eValuation XAI (VXAI) framework.}
\label{table:xai}
\end{table*}

\subsection{Evaluating Understanding and Trust}
A second line of research has investigated embedding ToM-like reasoning directly within robots and assessing its impact on trust, helpfulness, and mutual understanding. 
Some studies have focused on evaluating user perception, revealing that robots equipped with ToM capabilities are perceived more positively \cite{Mou}, particularly when they provide assistance aligned with users’ goals \cite{Cantucci}. 
Similarly, robots that reason about human beliefs are generally considered more helpful and socially competent \cite{Shvo}, and are also regarded as more trustworthy \cite{Angelopoulos}. 
At the same time, when providing explanations, robots may fail to enhance user understanding or improve decision-making, as not all explanations are equally effective \cite{Yuan}.
In contrast, approaches that implement multiple levels of explanation have been shown to improve user comprehension and the interaction \cite{Kerzel}.
Although these studies evaluate human–AI collaboration and occasionally describe their work as XAI, none assess it using XAI-specific criteria. 
Moreover, none have explicitly integrated ToM with XAI, highlighting a gap that our work addresses.

\section{A Comprehensive Evaluation of ToM}
While the field of ToM primarily claims to enhance user understanding, trust, and, more broadly, human–AI collaboration, these claims are often not systematically evaluated. 
This lack of evaluation stems from the fact that, if ToM purports to provide explanations for users, it should be assessed using the same criteria applied in XAI.
Indeed, these objectives align closely with those of XAI, which aims to design AI systems that are interpretable and comprehensible to humans \cite{Rong}. 
We therefore propose to systematically evaluate ToM in line with the rigor of its claims using an XAI evaluation framework. 

From this perspective, we examined whether existing ToM studies could be assessed using an XAI framework, positioning ToM itself as a framework for explanations.
To this end, we evaluated state-of-the-art ToM studies using the seven desiderata defined in the eValuation XAI (VXAI) framework \cite{dembinsky}, which integrates the principal evaluation criteria identified in recent reviews and provides a comprehensive and systematic approach to XAI evaluation.
We evaluated the studies based on whether they explicitly addressed each of the VXAI desiderata in their work. The criteria and detailed definitions of the desiderata are provided in Appendix\ref{appendix:desiderata}.

As shown in Table \ref{table:xai}, all of the ToM studies satisfy the Parsimony and Plausibility desiderata, indicating that they conducted user-centered experiments and assessed whether the explanations provided were perceived as believable.
However, only two studies meet the Continuity and Consistency criteria, reflecting that many experiments either did not report the number of participants or involved fewer than 100 participants. 
This poses a limitation for scaling these findings to real-world applications and for reproducibility.
None of the studies satisfies the Coverage desideratum, as none reported the number of successful versus unsuccessful interactions. 
Similarly, the Fidelity desideratum has not been addressed, indicating that none of the studies examined the internal reasoning process of the model. 
Consequently, it remains unclear whether the explanations accurately reflect the model’s behavior, which raises the risk of misleading users.

Together, these findings suggest that while ToM provides a valuable framework for user-centered evaluation, current studies lack rigorous XAI assessments.
This work seeks to address that gap by integrating ToM within an XAI framework, where fidelity serves as a central criterion for evaluating the alignment between explanations and the model’s actual reasoning.

\section{ToM as a User-Centered XAI Solution} 
Effective XAI should provide explanations that are both faithful to the system’s reasoning and understandable to users \cite{Rong}. 
Since existing methods often overlook user-centered design \cite{Rong}, we propose an integrated approach that combines ToM’s user focus with model-centered XAI techniques.
This integration enables explanations that are simultaneously interpretable for users and faithful to the model’s internal reasoning.
While some existing ToM integrations have proposed using Bayesian reinforcement learning to model user behavior, future work could explore the use of behavior trees or explainable reinforcement learning (XRL) within ToM-based systems.
Such approaches may support adaptive reasoning while further enhancing the fidelity of explanations.

\appendix
\section{Appendix 1}
\label{appendix:desiderata}
The seven desiderata proposed in VXAI \cite{dembinsky} can be described as follows:

\begin{itemize}
    \item \textbf{Parsimony}: Explanations should remain succinct and avoid unnecessary complexity to improve human understanding.
    \item \textbf{Plausibility}: The explanation must correspond to human logic and intuition, making it believable and relatable.
    \item \textbf{Coverage}: Indicates whether an explanation can be generated for each relevant input or output case.
    \item \textbf{Fidelity}: The explanation should truthfully mirror the underlying decision-making process of the model.
    \item \textbf{Continuity}: Measures the robustness of explanations when minor variations are introduced into the input data.
    \item \textbf{Consistency}: Ensures that explanations are coherent and reproducible for identical or comparable instances.
    \item \textbf{Efficiency}: Reflects the practicality of the explanatory approach in terms of computational cost and its general applicability across different models or domains.
\end{itemize}

For evaluation purposes, the following mapping criteria are applied:
\begin{itemize}
    \item When a human evaluation is conducted, the \textit{parsimony} and \textit{plausibility} desiderata are considered assessed.
    \item When the number of successful versus failed interactions is reported, the \textit{coverage} desideratum is considered assessed.
    \item When the model’s internal reasoning process is examined, the \textit{fidelity} desideratum is considered addressed.
    \item If the study reports a minimum of 100 human participants, both \textit{continuity} and \textit{consistency} are deemed evaluated.
    \item If computational implementation details are provided, \textit{efficiency} is regarded as evaluated.
\end{itemize}

\section*{Acknowledgments}
The authors gratefully acknowledge funding from Horizon Europe under the MSCA grant agreement No 101168792 (SWEET) and No 101072488 (TRAIL).

\bibliography{aaai2026}

\end{document}